\documentclass[preprint,12pt,authoryear]{elsarticle}

\usepackage{natbib}
\usepackage{verbatim}
\usepackage{caption}
\usepackage{subfigure}
\usepackage{color}

\usepackage{rotating}

\usepackage{url}
\begin{document}

\title{Identifying Sub-Phenotypes of Acute Kidney Injury using Structured and Unstructured Electronic Health Record Data with Memory Networks} 

\author{Zhenxing Xu$^1$, Jingyuan Chou$^1$, Xi Sheryl Zhang$^1$, Yuan Luo$^2$, Tamara Isakova$^2$, Prakash Adekkanattu$^1$, Jessica S. Ancker$^1$, Guoqian Jiang$^3$, Richard C. Kiefer$^3$, Jennifer A. Pacheco$^2$, Luke V. Rasmussen$^2$,  Jyotishman Pathak$^1$,  Fei Wang$^{1,*}$\\
\vspace{1em}
 $^1$ Weill Cornell Medicine, New York, NY, USA\\
 $^2$ Northwestern University, Chicago, IL, USA\\ 
 $^3$ Mayo Clinic, Rochester, MN, USA\\ 
 $*$ Corresponding author. Email: few2001@med.cornell.edu }

\begin{abstract}
Acute Kidney Injury (AKI) is a common clinical syndrome characterized by the rapid loss of kidney excretory function, which aggravates the clinical severity of other diseases in a large number of hospitalized patients. Accurate early prediction of AKI can enable in-time interventions and treatments. However, AKI is highly heterogeneous, thus identification of AKI sub-phenotypes can lead to an improved understanding of the disease pathophysiology and development of more targeted clinical interventions. This study used a memory network-based deep learning approach to discover AKI sub-phenotypes using structured and unstructured electronic health record (EHR) data of patients before AKI diagnosis. We leveraged a real world critical care EHR corpus including 37,486 ICU stays. Our approach identified three distinct sub-phenotypes: sub-phenotype I is with an average age of 63.03$ \pm 17.25 $ years, and is characterized by mild loss of kidney excretory function (Serum Creatinine (SCr) $1.55\pm 0.34$ mg/dL, estimated Glomerular Filtration Rate Test (eGFR) $107.65\pm 54.98$ mL/min/1.73$m^2$). These patients are more likely to develop stage I AKI. Sub-phenotype II is with average age 66.81$ \pm 10.43 $ years, and was characterized by severe loss of kidney excretory function (SCr $1.96\pm 0.49$ mg/dL, eGFR $82.19\pm 55.92$ mL/min/1.73$m^2$). These patients are more likely to develop stage III AKI. Sub-phenotype III is with average age 65.07$ \pm 11.32 $ years, and was characterized moderate loss of kidney excretory function and thus more likely to develop stage II AKI (SCr $1.69\pm 0.32$ mg/dL, eGFR $93.97\pm 56.53$ mL/min/1.73$m^2$). Both SCr and eGFR are significantly different across the three sub-phenotypes with statistical testing plus postdoc analysis, and the conclusion still holds after age adjustment.

\end{abstract}

\maketitle

\section{Introduction}
Acute Kidney Injury (AKI) is a critical clinical event characterized by a sudden decrease in kidney function, which affects about 15\% of all hospitalizations and more than 50\% of patients in intensive care unit (ICU) \citep{cheng2017predicting}. It is a complex condition, which rarely possesses an unique and obvious pathophysiology \citep{makris2016acute}. Patients with AKI usually present a complex etiology, which involves different causes and complicates the clinical manifestations and treatments \citep{tomavsev2019clinically}. For example, sepsis, ischemia, and the nephrotoxicity of drugs frequently co-occur in patients with AKI \citep{makris2016acute,chertow2005acute}. Moreover, AKI is a progressive kidney function disorder and has multiple stages in terms of severity, which is associated with a broad spectrum of clinical factors including creatinine, chloride, and hemoglobin measurements, as well as heart and respiration rates. Therefore, identification of AKI subtypes can lead to an improved understanding of the underlying disease etiology and the future development of more targeted interventions and therapies. However, due to the complexity and heterogeneity of AKI, it is challenging to define AKI sub-phenotypes accurately based on clinical knowledge. 

In recent years, due to wider availability of electronic health record (EHR) data, researchers have developed data-driven approaches to identify disease sub-phenotypes using EHR data \citep{fereshtehnejad2017clinical, maglanoc2019data}. From a data-driven perspective, patient sub-phenotyping is essentially a clustering problem \citep{zhang2019data,fereshtehnejad2017clinical}, where  patients in the same "sub-phenotype cluster" tend to be more similar to each other based on disease manifestations derived from their EHRs. Generally speaking, there are three steps for the task of data-driven discovery of sub-phenotypes using EHRs:
\begin{itemize}
\item {\em Representation}. For each patient in the dataset, we need to first construct an appropriate representation (e.g., vectors \citep{sun2012combining}, matrices \citep{wang2013framework}, tensors \citep{luo2016tensor} or sequences \citep{baytas2017patient}) of their information derived from the EHRs. 

\item {\em Clustering}. Next, we need to develop or adopt existing clustering algorithms (e.g., hierarchical agglomerative clustering \citep{fereshtehnejad2017clinical} or K-means \citep{zhang2019data}) based on the patient representations derived from previous step to acquire the patient clusters. Each cluster corresponds to an unique sub-phenotype.

\item {\em Interpretation}. After the patient clusters are obtained, we need to derive a clinical characterization for each of them (usually by identification of features that are discriminative across the different clusters through statistical testing \citep{fereshtehnejad2017clinical,zhang2019data}). This step allows us to appropriately interpret the computationally derived clusters.
\end{itemize}

Several prior studies have demonstrated the utility of data-driven sub-phenotyping with patient EHRs. For example, Ho {\em et al.} \citep{ho2014limestone} developed a tensor factorization based phenotyping approach that was capable of exploring the interactions among different modalities in the EHR data (e.g., diagnosis and medication). Zhou {\em et al.} \citep{zhou2014micro} proposed a matrix factorization algorithm to first reduce the dimensionality of the event space in the EHRs and then derive the the patient phenotypes in the low-dimensional space. Baytas \textit{et al.} \citep{baytas2017patient} further developed a deep learning approach to exploit the event temporalities in patient EHR. While promising, these prior efforts were limited to analyzing only structured EHR data. More recent work has also investigated unstructured EHR data to computationally derive sub-phenotypes \citep{halpern2016electronic,mccoy2018high}. In reality, both structured and unstructured EHR data contain important information about the patients, and therefore, an ideal solution to derive sub-phenotype should integrate all available EHR data. To this end, Pivovarov {\em et al.} \citep{pivovarov2015learning} proposed a mixture topic modeling approach for large-scale discovery of computational models of disease or phenotypes. However, their method does not explore the temporality between the clinical events documented in both structured and unstructured EHR data. 


In this study, we propose a deep learning model architecture to identify sub-phenotypes of AKI using longitudinal structured and unstructured EHR data. In particular, this study contains two goals. The first goal is to obtain effective representations that can accurately predict the incidence of AKI cases, and the second goal is to identify AKI subtypes on the obtained representations. More concretely, our approach is composed of three steps:
\begin{itemize}
\item In the representation step, we choose memory network (MN) \citep{sukhbaatar2015end} as the backbone of our model, wherein the structured event sequences are inserted into the memory network and the unstructured clinical note series are transformed into a vector through a  hierarchical Long Short Term Memory (HieLSTM) model. The vector is then combined with useful information extracted from the memory network to construct a vector representation for each patient. We optimize the model parameters such that the final representation vector can lead to the best prediction performance of a future AKI risk. 

\item In the clustering step, we first used the student t-distributed Stochastic Neighbor Embedding (t-SNE)\citep{maaten2008visualizing,van2009dimensionality} to project the patient vectors obtained into a two-dimensional space such that the cluster structure can be inspected visually, and then performed K-means clustering to obtain the patient clusters.

\item In the interpretation step, we performed statistical testing and manual evaluation to identify the features that are discriminant across the clusters, and used those features for defining disease sub-phenotypes.
\end{itemize}

We applied this approach to detect AKI sub-phenotypes using the EHRs from the MIMIC III data set \citep{johnson2016mimic}, where three distinct sub-phenotypes were identified that align with the different stages of AKI. In the following we introduce our study in detail.

\section{Methods}

\subsection{Data Set, AKI Case Definition, and Patient Features}
\label{sec:dataset}

The EHR data used in this study is from the Medical Information Mart for Intensive Care III (MIMIC-III) database \citep{johnson2016mimic}, which is a de-identified and publicly available data set. It contains approximately sixty thousand admissions of patients who stayed in critical care units of the Beth Israel Deaconess Medical Center between 2001 and 2012. The patient information contained in this database includes patient demographics, vital signs, laboratory test results, procedures, medications, clinical notes, imaging reports, and patient mortality.

\underline{\emph{AKI Case Definition}}: There are four commonly used AKI diagnostic criteria: the Risk-Injury-Failure-Loss-End (RIFLE) criteria \citep{bellomo2004acute}, then pediatric RIFLE (pRIFLE) criteria \citep{akcan2007modified}, the Acute Kidney Injury Network (AKIN) criteria \citep{pickering2009gfr}, and the Kidney Disease: Improving Global Outcomes (KDIGO) criteria \citep{kellum2012kidney}. KDIGO is the latest and it unified the previous criteria in 2012, which improves the sensitivity of AKI diagnosis and has been widely used by researchers \citep{makris2016acute,li2018early}. In this study, we employ the KDIGO criteria to define AKI cases as follows: (1) Increase in Serum Creatinine (Scr) by $\geq $ 0.3 mg/dl ($\geq$26.5 μmol/l) within 48h; or (2) Increase in SCr to $\geq$1.5 times from the baseline, which is known or presumed to have occurred within the prior 7 days; or (3) Urine volume $<$ 0.5 ml/kg/h for 6h. Moreover, as we will interpret the data-driven sub-phenotypes based on AKI stages, we will consider the following criteria from KDIGO for AKI staging:

Stage 1: SCr is 1.5-1.9 times of baseline, which is known or presumed to have occurred within the prior 7 days; or no less than 0.3 mg/dL (26.5 mol/L) absolute increase in SCr within 48h; or urine volume is less than 0.5 mL/kg/h for 6-12 hours.

Stage 2: SCr is no less than 2.0-2.9 times of baseline, which is known or presumed to have occurred within the prior 7 days; or urine volume is less than 0.5 mL/kg/h for no less than 12 hours.

Stage 3: SCr is no less than 3.0 times from baseline, which is known or presumed to have occurred within the prior 7 days; or increase in SCr is no less than 4.0 mg/dL (353.6 mol/L) within 48h; or initiation of renal replacement therapy; or urine volume is less than 0.3 mL/kg/h for no less than 24 hours; or anuria for no less than 12 hours; or renal replacement therapy required.

\underline{\emph{Patient Features}}: 
We extracted several groups of features from MIMIC-III as follows\footnote{Additional details about features can referred \citep{luo2016predicting} and be found at \url{https://github.com/xuzhenxing2018/amia/blob/master/features_name.xlsx}}.
\begin{itemize}
\item Demographics: Gender, age and ethnicity.

\item Medications: Medications that were administered from the patients' ICU admission until prediction time. Based on literature analysis and the KDIGO criteria, we mainly consider the following categories: diuretics, Non-steroidal anti-inflammatory drugs (NSAID), radiocontrast agents, and angiotensin.

\item Comorbidities: Comorbidities that the patients already have. Based on literature analysis and the KDIGO criteria, we mainly consider the following categories: congestive heart failure, peripheral vascular, hypertension, diabetes, liver disease, myocardial infarction (MI), coronary artery disease (CAD), cirrhosis, and jaundice.

\item Chart-events: Vital signs measured at the bedside. Based on literature analysis and the KDIGO criteria, we mainly consider diastolic blood pressure (DiasBP), glucose, heart rate, mean arterial blood pressure (MeanBP), respiration rate,  blood oxygen saturation level (SpO2), systolic blood pressure (SysBP), and temperature.

\item Lab-events: Laboratory test results performed prior to and during the ICU stay. Based on literature analysis and the KDIGO criteria, we consider the following tests: bicarbonate, blood urea nitrogen (BUN), calcium, chloride, creatinine, hemoglobin, international normalized ratio (INR), platelet, potassium,  prothrombin time (PT), partial thromboplastin time (PTT), white blood count (WBC), the average of urine output, and the minimum value of estimated glomerular filtration rate (eGFR) that is computed by MDRD (Modification of Diet in Renal Disease) 4-Variable Equation including age, sex, ethnicity, and serum creatinine \citep{levey2006using}. 
\end{itemize}


\subsection{Experimental Setting and Data Pre-Processing}

\begin{figure*}
\centering
\includegraphics[scale=0.1]{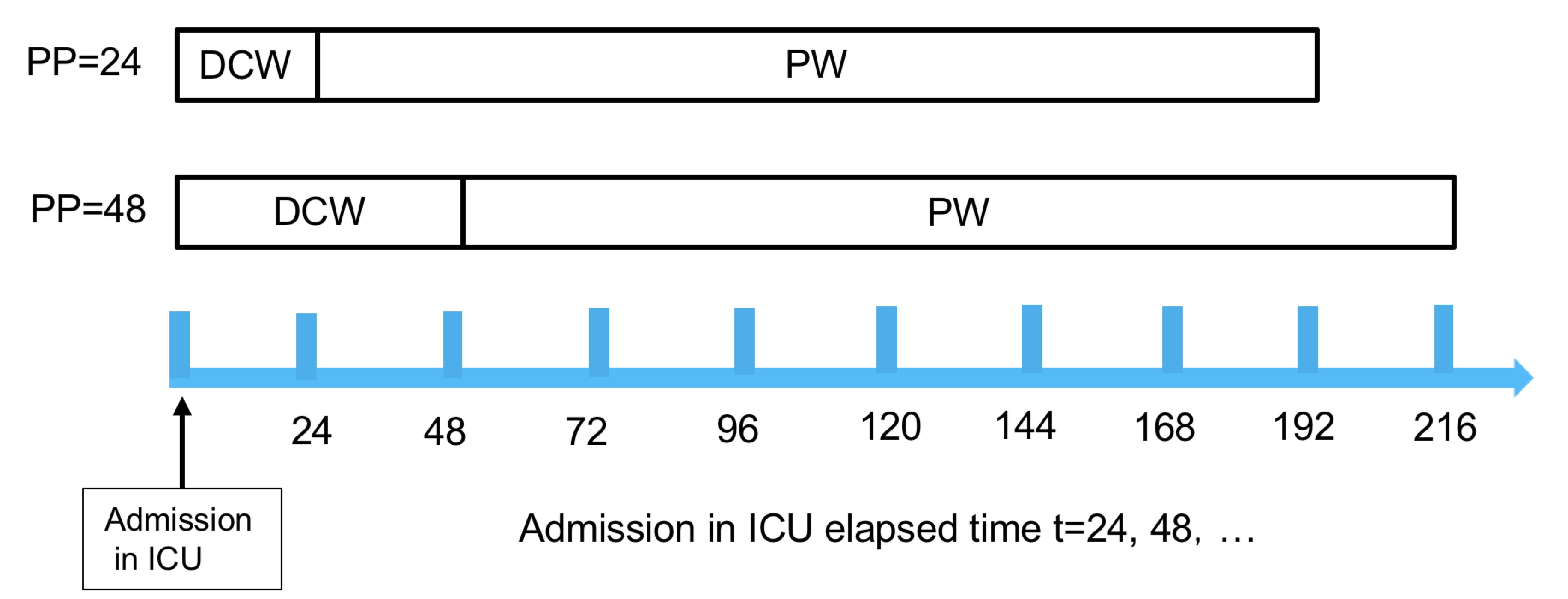}
\caption{An illustration of AKI prediction in our experimental setting. PP: prediction point; DCW: data collection window; PW: prediction window.}
\label{fig:Experimental_setting}
\end{figure*}

As discussed in the Introduction, the first step of data-driven sub-phenotyping is to learn an appropriate representation for each patient. In our approach, we expect such representation can effectively predict future AKI risk. In particular, because MIMIC-III mainly contains ICU admission data for the patients in critical care, and the same individual might have multiple ICU stays, we define AKI cases and controls based on the records in each ICU stay. More concretely, let $t_1$ be the elapsed time (in hours) after the patient was admitted to ICU, from which we will extract the patient records to train our model. Thus $t_1$ is also referred to as observation window. The AKI case/control label is defined on the records within $t_2$ after $t_1$. Thus $t_2$ is also referred to as prediction window. In this study,  $t_1$ is set to 24 hours (or 48 hours) and $t_2$ is set to 7 days. The illustration of experimental setting is shown in Figure~\ref{fig:Experimental_setting}. We choose 7-days as the prediction window according to  the definition of AKI by KDIGO criteria. Related study would suggest 7 days as the cut-off for diagnosing AKI \citep{schneider2017aki}. While the 0.3 mg/dl increase for urine is within 48 hours, the definition of  stage 1, 2, 3 are 1.5 times  of  baseline, 2 times  of  baseline, 3 times  of  baseline within 7 days, respectively. Technically, when a patient enters the ICU unit they might not have had any kidney injury but the thought is that many of them are at risk due to low blood pressure, inflammation, or sepsis. We exclude patients who were admitted with AKI in the ICU and had AKI in the first 24 hours (or 48 hours) in the ICU. We also exclude patients whose SCr and urine data are missing from the prediction window.

For data pre-processing, we first categorize the observation window into a set of equal-length sub-windows, and we set the sub-window length to 2 hours in our study. The values for each variable are averaged over each sub-window. We impute the missing values by the variable mean and further uniformly scale the values for each variable into the region between 0 and 1. The discrete variables (e.g., medication and comorbidities) are encoded as zero-one multi-hot vectors. For unstructured data, each clinical note is modeled as a sequence of words. The medical term dictionary was constructed based on our prior work \citep{li2018early}, which was further leveraged to filter the clinical notes. In this study, each piece of clinical notes is modeled as a sequence of words, which can keep important medical term words and some corresponding context words. These context words can provide some useful information for modelling each piece of clinical notes when using some deep learning based methods. From these notes, we can obtain some evidence and relationships in terms of the progress of different diseases. Note that, we do not use all words in notes, because the researchers removed some words when building medical term dictionary, e.g., the words with document-frequency under 100 were removed to reduce noise. More details in terms of the process of building medical term dictionary can be referred \citep{li2018early}. Note that each ICU stay typically contains multiple clinical notes. Based on these aforementioned data pre-processing steps, we end up with a total of 37,486 ICU stays including 7,657 AKI cases from 6,933 unique patients, and 29,829 controls from 24,366 unique patients.

\subsection{The Proposed Model}



\begin{figure*}
\centering
\includegraphics[scale=0.55]{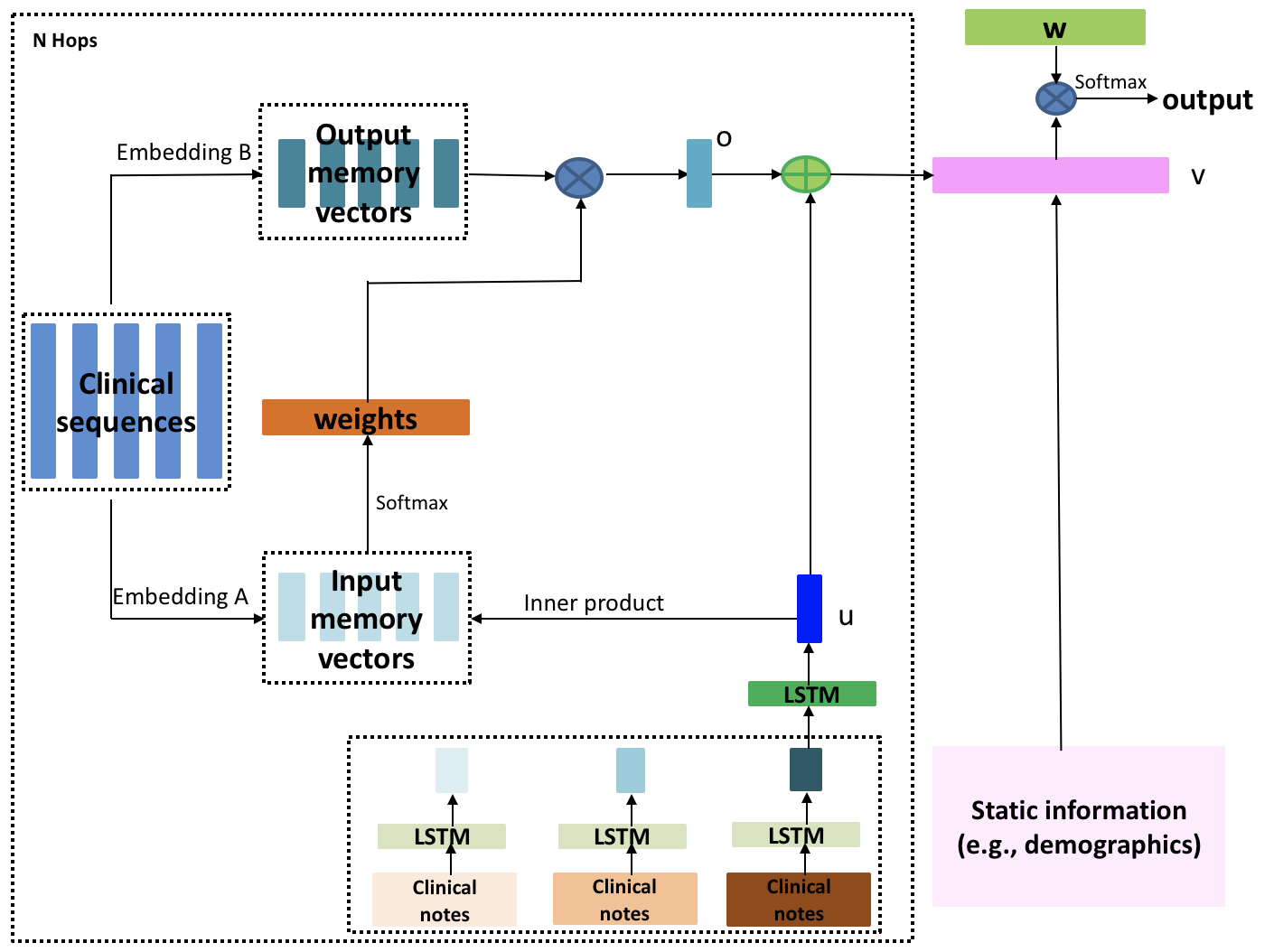}
\caption{An illustration of the proposed framework. An ICU stay representation $v$ is derived by integrating structured and unstructured EHR data. Time-dependent structured EHR data (clinical sequences) is push into the memory by using embedding matrices $A$ and $B$. Unstructured EHR data (clinical notes) is represented as a vector $u$ by using hierarchical LSTM, which is combined with input memory vectors to retrieve important information from output memory vectors to form a vector $o$. Static information is integrated with $o$ and $u$ to form an ICU stay representative vector $v$. }
\label{fig:framework}
\end{figure*}

As stated in the Introduction, there are three key steps in our model. The first step is to construct effective representation vectors. The overall architecture of our model is shown in Figure~\ref{fig:framework}, where we use memory networks (MN) \citep{weston2014memory,sukhbaatar2015end} as the backbone. The general assumption here is that the AKI risk during a specific ICU stay will be dependent on the information contained in the series of clinical notes (extracted from a hierarchical LSTM model) and structured fields (such as lab tests, vital signs, etc., which are stored in the memory bank) in the EHR before the prediction time point. We further elaborate our model in the following.

MN models have been demonstrated to be very effective in Question-Answering problems \citep{miller2016key, bordes2015large}. The general setup of an MN model includes a set of inputs $x$ that are to be stored in the memory (which contain information of potential answers), a query $q$, and an answer $a$. The MN model compares an input query $q$ with the information in the memory $x$ and retrieves useful information that is similar to $q$, and then combines the query with the retrieved memory information to predict what would be the right answer. Compared to conventional deep learning models such as Recurrent Neural Network (RNN) \citep{mikolov2010recurrent}, MN models can explicitly control parts of information in the memory that will be extracted, which is much more flexible and interpretable. Multiple previous studies demonstrated that MN based models can obtain better performance than several deep learning models (e.g., LSTM) \citep{weston2014memory, sukhbaatar2015end}. In addition, although several traditional deep learning models (e.g., LSTM) can integrate multi-model data \citep{rajkomar2018scalable}, the design of $“query”$, $“answer”$, and $“memory”$ in the MN models has a potential to integrate multi-model data to obtain good performance \citep{zhang2018integrative}.

In our setting, for each ICU stay, our goal is to learn patient representations that can best predict a future AKI risk. We propose to insert the structured EHR information within the observation window into the memory, and transform the series of clinical notes into a query vector. We then compare the query vector with the contents in the memory slots and extract relevant information which are combined with the query to make the AKI prediction. Two major operations in our model are reading and retrieving as discussed below.

\underline{\emph{Clinical sequences reading:}} In the MN model, suppose there is an input clinical sequence $s_j$, $j=1,...,t$, where $j$ and $t$ are the index of time slots and the number of time slots, respectively. Each item in $s_j$ is the value of some clinical variables extracted from the EHR at a certain timestamp. We mainly incorporate time-dependent structured information from charts events variables (e.g., heart rate, respiration rate, systolic blood pressure, and temperature) and lab test results (e.g., calcium, chloride, platelet, creatinine and  potassium) into the memory. We employ a fixed number of timestamps $t$ to define the memory size. An embedding matrix $A$ is used to transform input clinical sequences into continuous vectors $z_j=As_j$, and stored into the memory, which are regarded as a new input memory representation. Meanwhile, we use another embedding matrix $B$ to obtain output memory vectors $e_j=Bs_j$.

\underline{\emph{Memory retrieving: }}
The retrieving of memory representation is to find memory vectors from the embedding space. Since different clinical variables in a sequence can contribute differently when it comes to the representation learning for clinical notes, it is necessary to decide which vector to choose. Attentive weights are used here to make a soft combination of all memory vectors. In particular, the weights $\alpha$ are computed by a softmax function on the inner product of the input memory $z_i$ and the learned query vector $u$ from the clinical notes:
\begin{equation}
\alpha_i = Softmax(u^Tz_i)
\end{equation}
where $Softmax(r_i) = e^{r_i}/\sum_j e^{r_j}$. Defined in this way $\alpha$ is a probability vector over the inputs. Once these weights are obtained, we use these weights and output memory vectors $e_i$ to obtain a vector $o$ as 
\begin{equation}
o = \sum_i \alpha_ie_i,
\end{equation}
which is combined with $u$ to form a integrated vector $v$ to represent the integrated information within the observation window of an ICU stay.

Generally speaking, such a memory mechanism allows the network to read the input sequences multiple times to update the memory contents at each step and then make a final output. The architecture of multiple layers can be used to collect the information from the memory iteratively and cumulatively for learning the presentations of patients \citep{rush2015neural,xu2015show}. In particular, suppose there are $L$ layer memories for $L$ hop operations, the output feature map at the $l-th$ hop can be rewritten as
\begin{equation}
u^{l+1}=Hu^l + o^l,  (l = 1,...,L)
\end{equation}
where $H$ is a linear mapping and can be beneficial to the iterative updating of $u$. In addition, a layer-wise updating strategy for input and output memory vectors at multiple hops are used, which maintains the same embedding as $A^1=...=A^L$ and $B^1=...=B^L$. More details in terms of the process of using multiple layers in  memory networks can be referred \citep{sukhbaatar2015end}. 

To transform the clinical note series into a vector representation $u$, we use a two-layer hierarchical LSTM (HieLSTM) model. In particular, each ICU stay in the MIMIC dataset comprises multiple clinical notes with timestamps. The bottom layer LSTM is built on each specific clinical note whose word sequence is served as the input. The embedding vector in the hidden layer of the last word is the output. By concatenating the output vectors for all clinical notes in each ICU stay according to their timestamps we obtain the input of the top layer LSTM, and the embedding vector in the hidden layer at the last timestamp will be used as the query. 

We compute the probability distribution over the binary class by
\begin{equation}
P = Softmax(w^Tv)
\end{equation}
where $w$ is the coefficient vector and $v$ is the vector combined with memory output $o$ and query $u$. During training, we use cross-entropy to compute the prediction loss
\begin{equation}
\mathcal{L} = \sum_i^N y_ilogp_i + (1-y_i)log(1-p_i)
\end{equation}
where $y_i$ denotes the label for the $i$-th ICU stay, and $N$ is the total number of ICU stays in training data set. 

Our model is implemented with tensorflow 1.7.0 \citep {abadi2016tensorflow} and trained on workstations with NVIDIA TESTA V100 GPUs with mini-batch Adam optimizer \citep{kingma2014adam,zhang2018integrative}.  

Based on prior research \citep {sukhbaatar2015end}, we set the memory size and the dimension of embedding as 12 and 128, respectively. For HieLSTM, the dimension of each hidden units in the bottom and top LSTMs are set as 200 and 128, respectively. The final dimension of representation of each ICU stay is 144. In addition, batch size, learning rate, and epoch number are set as 32, 0.01, and 10, respectively.

\subsection{Baseline Models}
\label{sec:baselinemodels}

 In order to evaluate the effectiveness of the proposed model, we compare its prediction performance with the performances of a set of baseline models. Specifically, we implemented Logistic Regression (LR)\citep{le1992ridge},  Random Forest (RF) \citep{breiman2001random}, and Gradient Boosting Decision Tree (GBDT) \citep{friedman2002stochastic} as the baseline models. To construct the input vector of these models, for time-dependent continuous variables (e.g., lab tests and chart-events) derived from structured EHR data, we calculate the statistics within the observation window including the first, last, average, minimum, maximum, slope and the count. For discrete variables (e.g., medication and comorbidities) derived from structured EHR data, we encode them as zero-one multi-hot vectors. Demographics data are also considered. In particular, gender and ethnicity are seen as discrete variables and age is considered as continuous variable. In this way, the structured EHR data within the observation window of each ICU stay is represented as a 147-dimension feature vector. For unstructured clinical notes, we construct a bag-of-words vector from the collection of clinical notes in each ICU stay. For the implementations of LR and RF, we use the Scikit-learn software library \citep{pedregosa2011scikit}. For the GBDT, we use the XGBoost software library \citep{chen2016xgboost}. In addition to these conventional machine learning models, we also implement LSTM and two-layer hierarchical LSTM for prediction based on pure structured and unstructured EHR information.
 

\section{Results and Discussions}

\subsection{Patient Representation Learning}

As introduced above, the first step of our pipeline is to learn an effective representation that can predict the risk of AKI. In experiment, we used the nested cross-validation \citep{statnikov2004comprehensive}, which an outer 5-cross-validation was used to split our dataset into training data and testing data, and an inner 5-cross-validation on training data was used for tuning the parameters.  Table~\ref{tbl:table_24hour} and Table~\ref{tbl:table_48hour} showed the performance (average $\pm$ standard deviation) of our approach together with the baseline models introduced in Section \ref{sec:baselinemodels} in terms of AUC(Area Under the Receiver Operating Characteristic Curve), precision, and recall (cutoff of 0.5). From these results, we observe that:
\begin{itemize}
\item	Compared to the performance from only unstructured data (i.e., clinical notes), incorporating both structured and unstructured EHR data can lead to an improved performance. One potential reason is that the structured data contains  information related to AKI such as lab tests and chart events.

\item	The combination of unstructured data and structured data improved the performance, which suggests that there is complementary information contained in both structured and unstructured EHR that is beneficial to the prediction. In particular, clinical notes on MIMIC-III dataset mainly contain several types of notes: (1) echo reports; (2) ECG reports; (3) radiology reports; (4) summary of hospital course of a patient;  (5) discharge summary; (6) diagnose descriptions; (7) patient past medical histories.  These information can provide multiple evidences and a comprehensive profile for patient current pathophysiologic conditions, for example, patient past medical histories (e.g., cardiac surgery history) can provide implications for patient conditions. The information from multiple kinds of reports may show some connections with AKI.

\item	Deep learning models (e.g., LSTM and HieLSTM) obtained better results than traditional classification algorithms in all three cases. One potential reason is that such models can capture the temporal dependencies among the structured events and unstructured clinical notes, which could be beneficial to the prediction of AKI risk.

\item	Our proposed methodology, which integrates both structured and unstructured information through a deep learning model architecture, yields the best performance.
\end{itemize}
Because the last layer before the output for our proposed model shown in Figure 1 is a simple fully-connected logistic regression, its superior performance really comes from the good patient representation vectors that serve as the inputs to the logistic regressor.

In addition, in order to investigate the importance of using clinical notes, we further examined important words by ranking ``feature\_importances\_'' in the GBDT over bag-of-words. The top 10 important words are ``lasix'', ``ci'', ``labile'', ``insulin'', ``wires'', ``cabg'', ``dilantin'', ``neuro'', ``jaundiced'', and ``cvp''. These words appear to be clinically meaningful. For example, the word ``lasix'' is one kind of diuretic, which have a potential to provide treatment for the retention that are caused by kidney dysfunction. The word ``cabg (Coronary artery bypass surgery)'' provide an implication since these patients have a potential to suffer from kidney injury post-operatively \citep{kertai2016platelet}.

\begin{table}
\centering
\small
\caption{The prediction performance of different methods based on structured and unstructured EHR data produced during 24 hours after admission} 
\label{tbl:table_24hour}
\begin{tabular}{c|c|c|c|c}
\hline\hline
Data & Methods & AUC & Precision & Recall\\
\hline\hline
      & LR & 0.6596$ \pm 0.0061 $ & 0.2897$ \pm 0.0122 $ & 0.5532$ \pm 0.0128 $ \\
Unstr & RF & 0.6645$ \pm 0.0082 $ & 0.2941$ \pm 0.013 $ & 0.5536$ \pm 0.0135 $ \\
      & GBDT & 0.6797$ \pm 0.0071 $ & 0.2978$ \pm 0.0135 $ & 0.5603$ \pm 0.0142 $ \\
      & HieLSTM & 0.6998$ \pm 0.0113 $ & 0.2992$ \pm 0.0142 $ & 0.5605$ \pm 0.0149 $ \\
\hline
      & LR & 0.6823$ \pm 0.0052 $ & 0.3115$ \pm 0.0142 $ & 0.5834$ \pm 0.0152 $ \\
Str   & RF & 0.6999$ \pm 0.007 $ & 0.3136$ \pm 0.0143 $ & 0.5836$ \pm 0.0152 $ \\
      & GBDT & 0.7021$ \pm 0.0071 $ & 0.3455$ \pm 0.0151 $ & 0.5956$ \pm 0.0154 $ \\
      & LSTM & 0.7099$ \pm 0.0091 $ & 0.3604$ \pm 0.016 $ & 0.5959$ \pm 0.0162 $ \\
\hline
      & LR & 0.7199$ \pm 0.0062 $ & 0.4112$ \pm 0.0131 $ & 0.5997$ \pm 0.014 $ \\
Unstr & RF & 0.7201$ \pm 0.0094 $ & 0.4222$ \pm 0.0132 $ & 0.5998$ \pm 0.0143 $ \\
+Str  & GBDT & 0.7319$ \pm 0.0083 $ & 0.4336$ \pm 0.0142 $ & 0.6009$ \pm 0.0145 $ \\
      & LSTM & 0.7431$ \pm 0.0095 $ & 0.4531$ \pm 0.0149 $ & 0.6101$ \pm 0.0152 $ \\
      & MN+HieLSTM & {\bf 0.7753$ \pm 0.0126 $} & {\bf 0.4994$ \pm 0.0155 $} & {\bf 0.6304$ \pm 0.0161 $} \\
\hline\hline
\end{tabular}
\end{table}

\begin{table}
\centering
\small
\caption{The prediction performance of different methods based on structured and unstructured EHR data produced during 48 hours after admission} 
\label{tbl:table_48hour}
\begin{tabular}{c|c|c|c|c}
\hline\hline
Data & Methods & AUC & Precision & Recall\\
\hline\hline
      & LR & 0.6603$ \pm 0.0059 $ & 0.2901$ \pm 0.0117 $ & 0.5568$ \pm 0.0119 $ \\
Unstr & RF & 0.6651$ \pm 0.0074 $ & 0.2952$ \pm 0.014 $ & 0.5598$ \pm 0.0136 $ \\
      & GBDT & 0.6802$ \pm 0.007 $ & 0.2983$ \pm 0.0141 $ & 0.5692$ \pm 0.0149 $ \\
      & HieLSTM & 0.7002$ \pm 0.0116 $ & 0.3001$ \pm 0.0144 $ & 0.5703$ \pm 0.0151 $ \\
\hline
      & LR & 0.6899$ \pm 0.0047 $ & 0.3126$ \pm 0.0142 $ & 0.5935$ \pm 0.0153 $ \\
Str   & RF & 0.7003$ \pm 0.006 $ & 0.3145$ \pm 0.0145 $ & 0.5991$ \pm 0.0152 $ \\
      & GBDT & 0.7091$ \pm 0.0069 $ & 0.3502$ \pm 0.0151 $ & 0.5999$ \pm 0.0155 $ \\
      & LSTM & 0.7133$ \pm 0.0088 $ & 0.3664$ \pm 0.0163 $ & 0.6002$ \pm 0.0163 $ \\
\hline
      & LR & 0.7211$ \pm 0.0058 $ & 0.4201$ \pm 0.0142 $ & 0.6089$ \pm 0.0148 $ \\
Unstr & RF & 0.7293$ \pm 0.0095 $ & 0.4289$ \pm 0.0135 $ & 0.6112$ \pm 0.0149 $ \\
+Str  & GBDT & 0.7345$ \pm 0.0084 $ & 0.4397$ \pm 0.0141 $ & 0.6119$ \pm 0.0152 $ \\
      & LSTM & 0.7505$ \pm 0.0093 $ & 0.4609$ \pm 0.0152 $ & 0.6198$ \pm 0.0159 $ \\
      & MN+HieLSTM & {\bf 0.7798$ \pm 0.0131 $} & {\bf 0.5011$ \pm 0.0162 $} & {\bf 0.6397$ \pm 0.0177 $} \\
\hline\hline
\end{tabular}
\end{table}

\subsection{Identifying AKI Sub-Phenotypes}

We take EHR data produced during 24 hours after admission (i.e., pp =24 in Figure~\ref{fig:Experimental_setting}) as an example to investigate the discovering of AKI sub-phenotypes and perform statistical analysis in the following experiments. In particular, after the patient representations were obtained, we first used student t-distributed Stochastic Neighbor Embedding (t-SNE) \citep{maaten2008visualizing,van2009dimensionality} to embed the patient data into a 2-dimensional space, such that the information can be visually inspected for identification of potential clusters. Then K-means clustering was performed in the 2-dimensional space to computationally identify the  clusters.  In our study, since the goal is to identify AKI sub-phenotypes, we perform clustering on the representation vectors of all the 7,657 AKI cases.

\begin{figure}[htbp]
    \centering
    \subfigure[]{
    	\includegraphics[width=1.65 in, height=1.655 in]{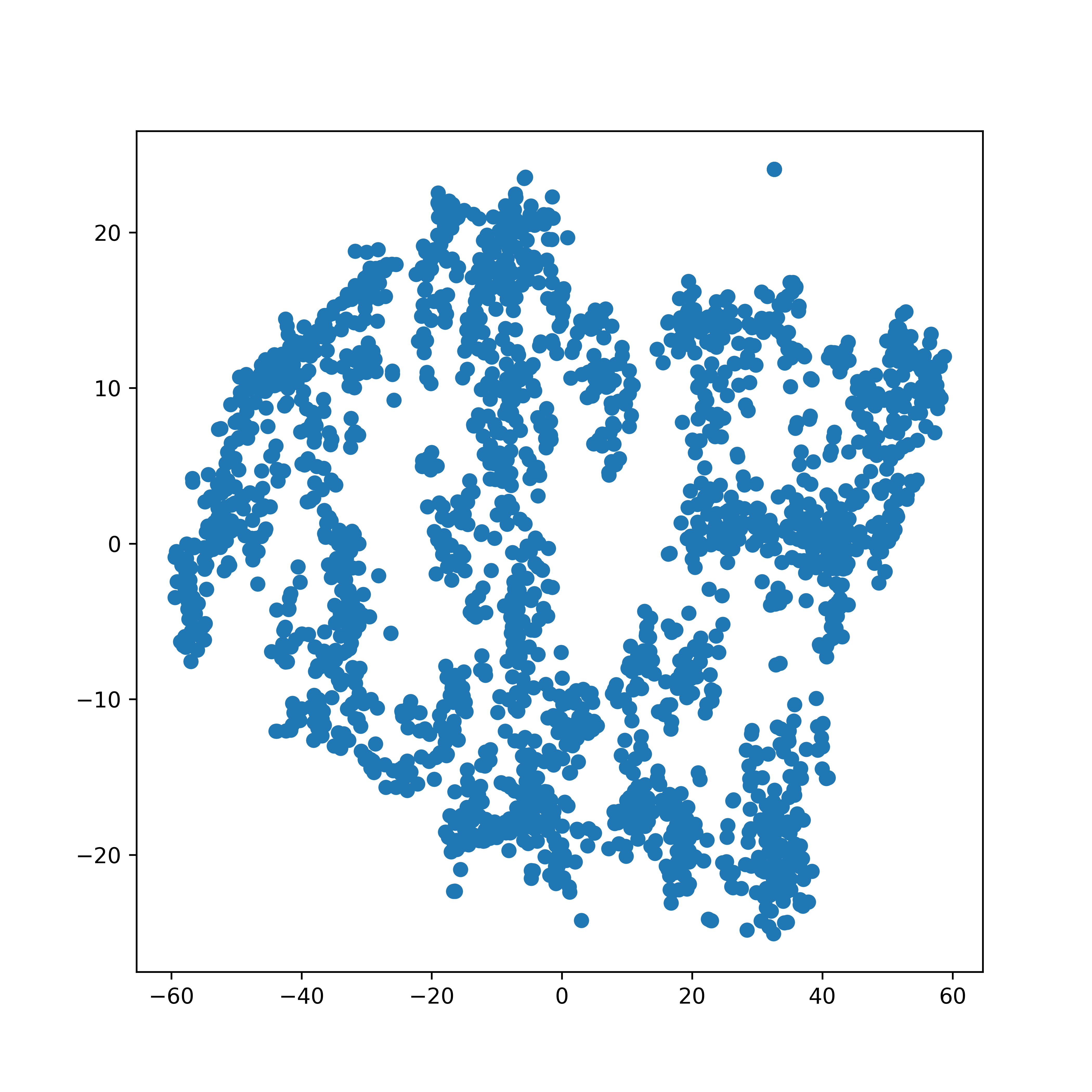}
    }
    \subfigure[]{
	\includegraphics[width=1.65 in, height=1.655 in]{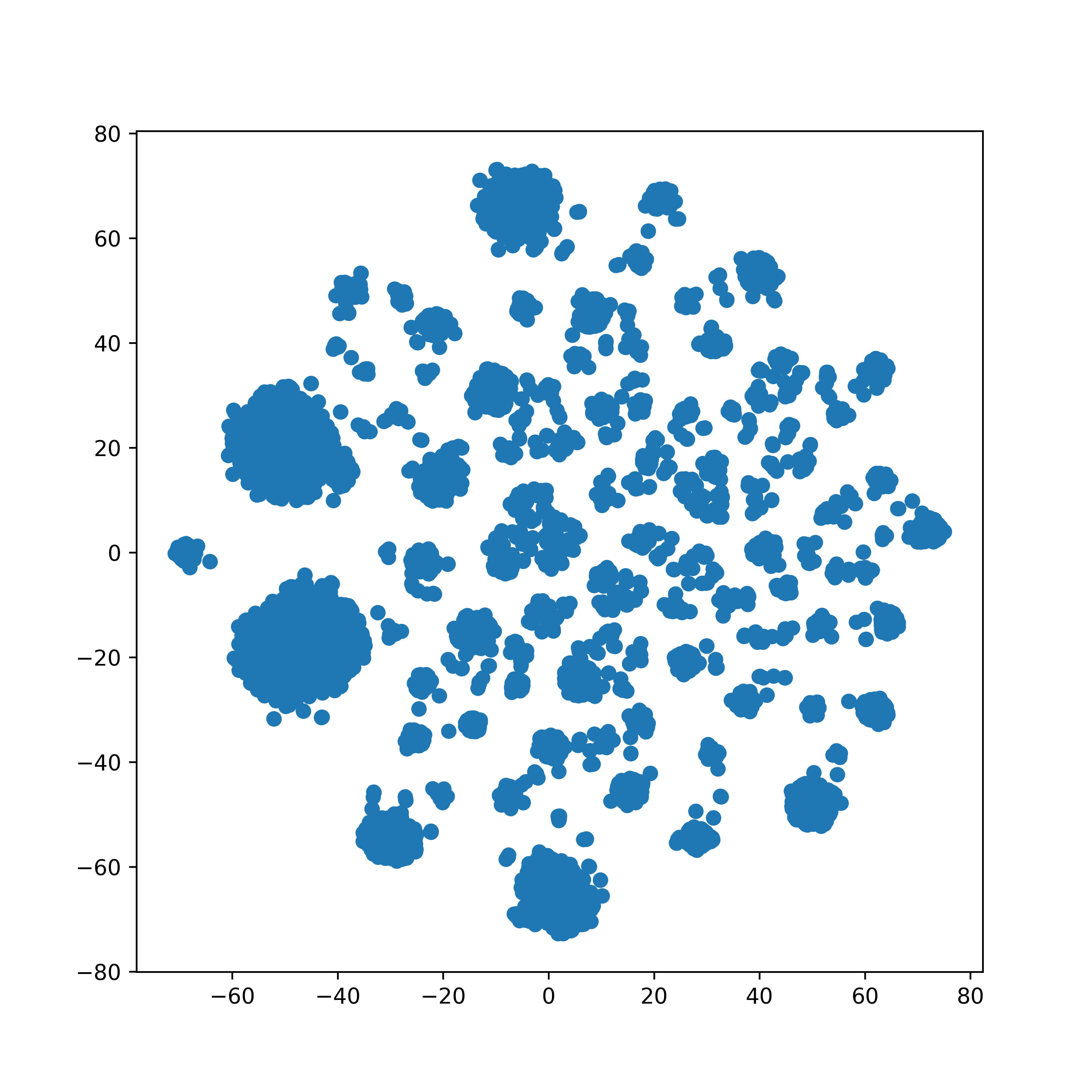}
    }
    \subfigure[]{
	\includegraphics[width=1.65 in , height=1.655 in]{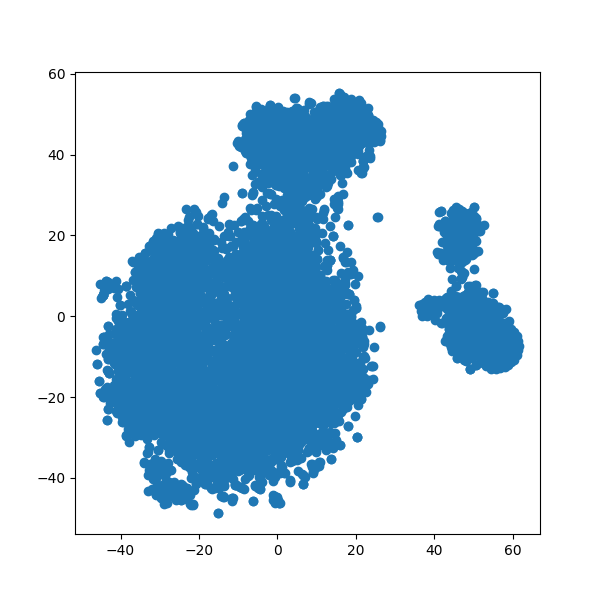}
    }
    \caption{The result of t-SNE: (a), (b), and (c) indicate the representation of patients is obtained by PCA, auto-encoder, and our method, respectively.}
    \label{FIG:t-SNE}
\end{figure}

\begin{figure}[htbp]
    \centering
    \subfigure[]{
        \includegraphics[width=1.65 in, height=1.655 in]{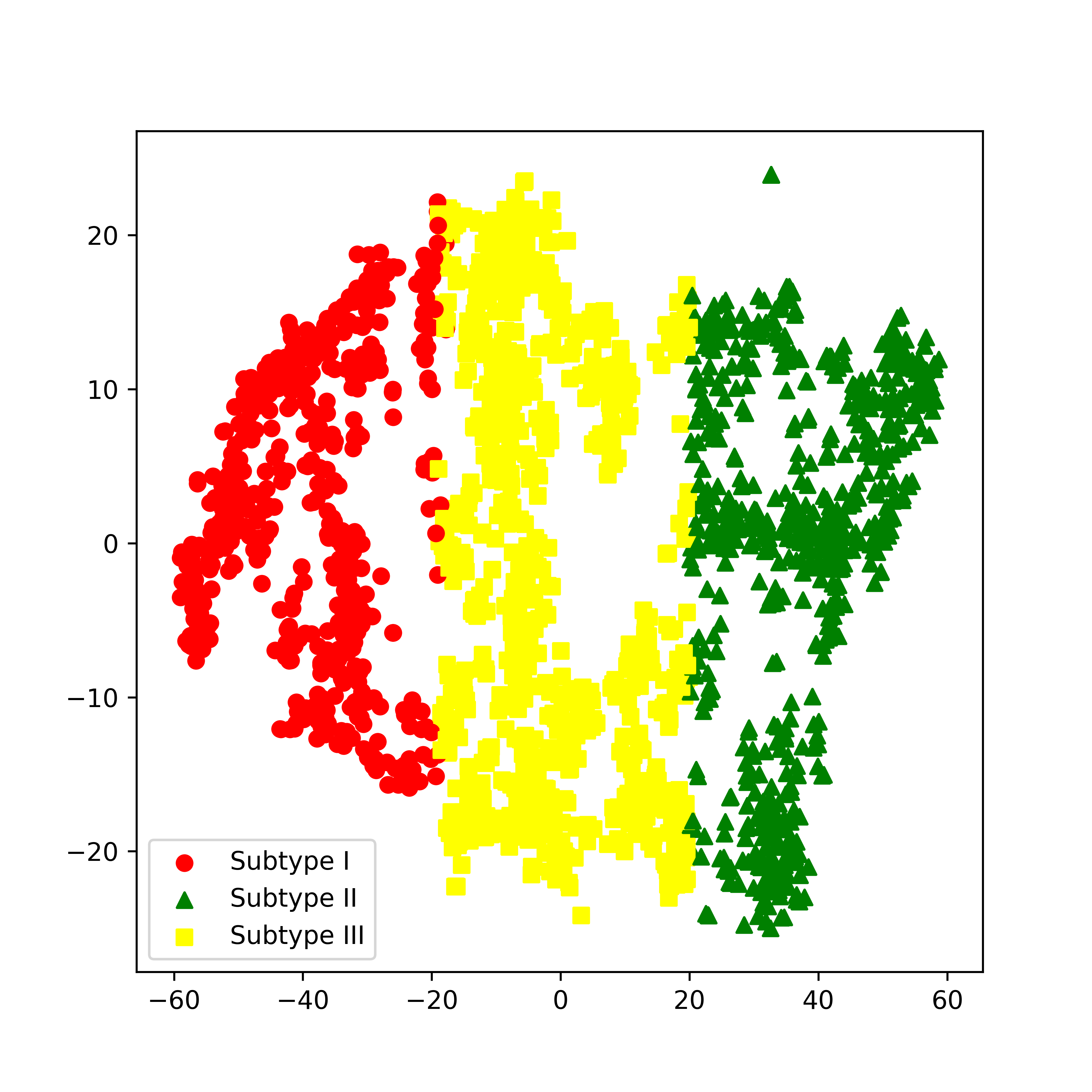}
    }
    \subfigure[]{
	\includegraphics[width=1.65 in,height=1.655 in]{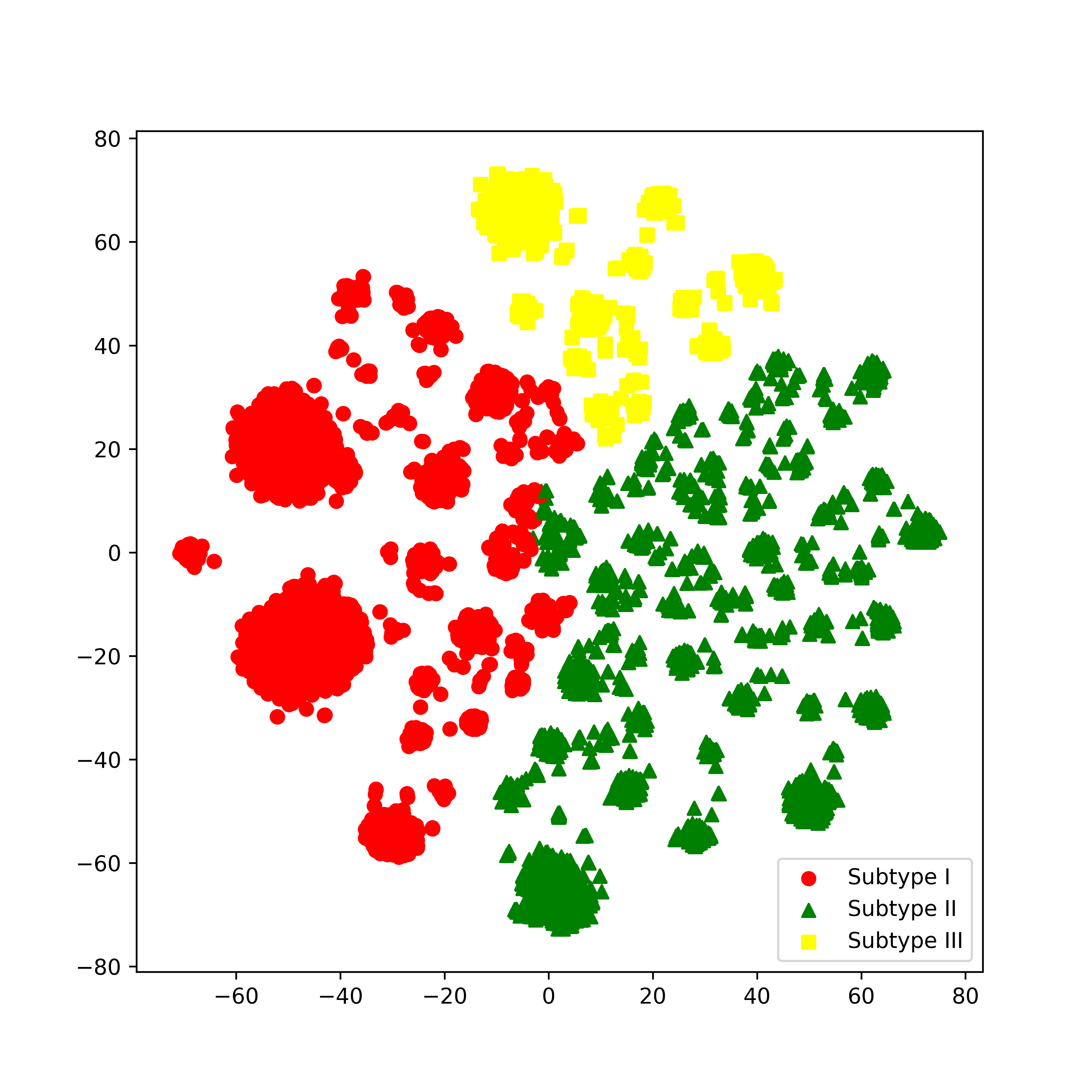}
    }
    \subfigure[]{
	\includegraphics[width=1.65 in, height=1.655 in]{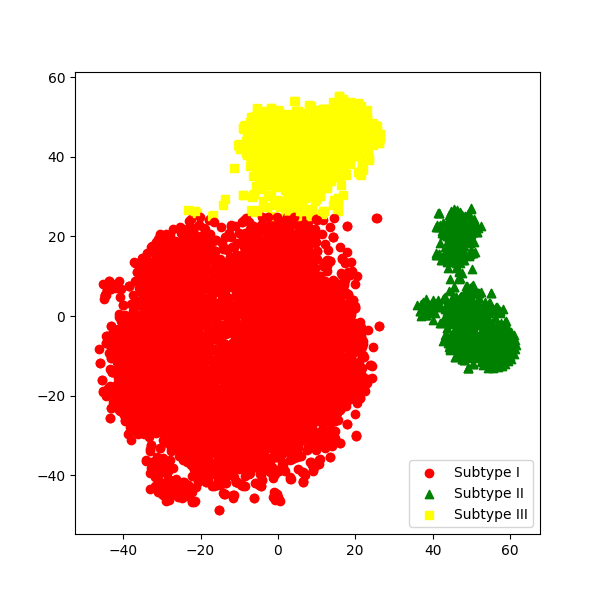}
    }
    \caption{The results of k-means clustering: (a), (b), and (c) indicate the representation of patients is obtained by PCA, auto-encoder, and our method, respectively.}
    \label{FIG: k-means results}
\end{figure}

The results of t-SNE embedding and clustering results are shown in Figure~\ref{FIG:t-SNE} and Figure~\ref{FIG: k-means results}. We consider PCA and auto-encoder methods as two baselines to compare the clustering results. Compared with Figure~\ref{FIG:t-SNE} (a)(b) and Figure~\ref{FIG: k-means results} (a)(b), the three sub-phenotype depicted in Figure~\ref{FIG:t-SNE} (c) and Figure~\ref{FIG: k-means results} (c) are much more salient, with a better separation in scatterplot (smaller intra-cluster distance and larger inter-cluster distance). The optimal number of clusters, which is three in this case, is determined by the Mcclain-Rao index \citep{mcclain1975clustisz}. The identified patient clusters are represented with different colors in Figure~\ref{FIG: k-means results}. In the following subsection, we discuss the interpretation of these computationally derived AKI sub-phenotypes.

\subsection{AKI Sub-Phenotype Interpretation}

In order to interpret the three sub-phenotypes, we performed statistical analyses on each of them to identify the patient features that are discriminative among them. In particular, we performed Chi-square test \citep{lowry2014concepts} and One-way ANOVA test \citep{mcdonald2009handbook} for normally distributed features, and non-parametric testing \citep{kruskal1952use} for features that are not normally distributed. Posthoc Tukey HSD test \citep{tukey1949comparing} was performed for multiplicity correction. Note that, the average of multiple measurements in terms of continuous variables within first 24-hour of admission are computed for statistical comparison between sub-phenotypes. The results are shown in Table~\ref{tbl:table_Discrete} and Table~\ref{tbl:table_Continuous}. From these tables, we can see that the following features are significantly different across the three subtypes: Glucose, Albumin, Diastolic Blood Pressure (DiasBP), Age, Lactate, Creatinine, Hemoglobin, Partial Thromboplastin Time (PTT), White Blood Count (WBC), Urine, and estimated Glomerular Filtration Rate (eGFR). After Age adjustment, the following features are still significantly different:
\begin{itemize}
    \item Glucose. Diabetes has been shown to be a big risk factor for kidney disease and there are lots of studies on diabetic kidney disease \citep{de2011temporal,seaquist1989familial,tuttle2014diabetic}. It has also been shown that ``AKI risk is most likely being increased in diabetic individuals" \citep{patschan2016acute}. 
    
    \item Albumin. Albumin is a protein made by the liver. When the kidney begins to fail, the albumin will leak to the urine and cause a low blood albumin. 
    
    \item Creatinine. Creatinine is a waste product from the normal breakdown of muscle tissue, which is filtered through the kidneys and excreted in urine. Creatinine level measures the kidney function.
    
    \item WBC. The white blood cells are cells fighting infections. A high WBC level may indicate problems like infection or inflammation. As inflammatory response plays a key role in the development of AKI, there are also studies on how WBC and AKI are related (e.g., \citep{han2014u}).
    
    \item Urine. This is the 24-hour urine volume test, which measures the kidney function.
    
    \item eGFR. This stands for the glomerular filtration rate estimated based on the creatinine test. It also measures the kidney function.
\end{itemize}
Therefore, all these identified feature variables are highly correlated with kidney function and AKI.

\begin{figure}
	\centering
		\includegraphics[scale=.37]{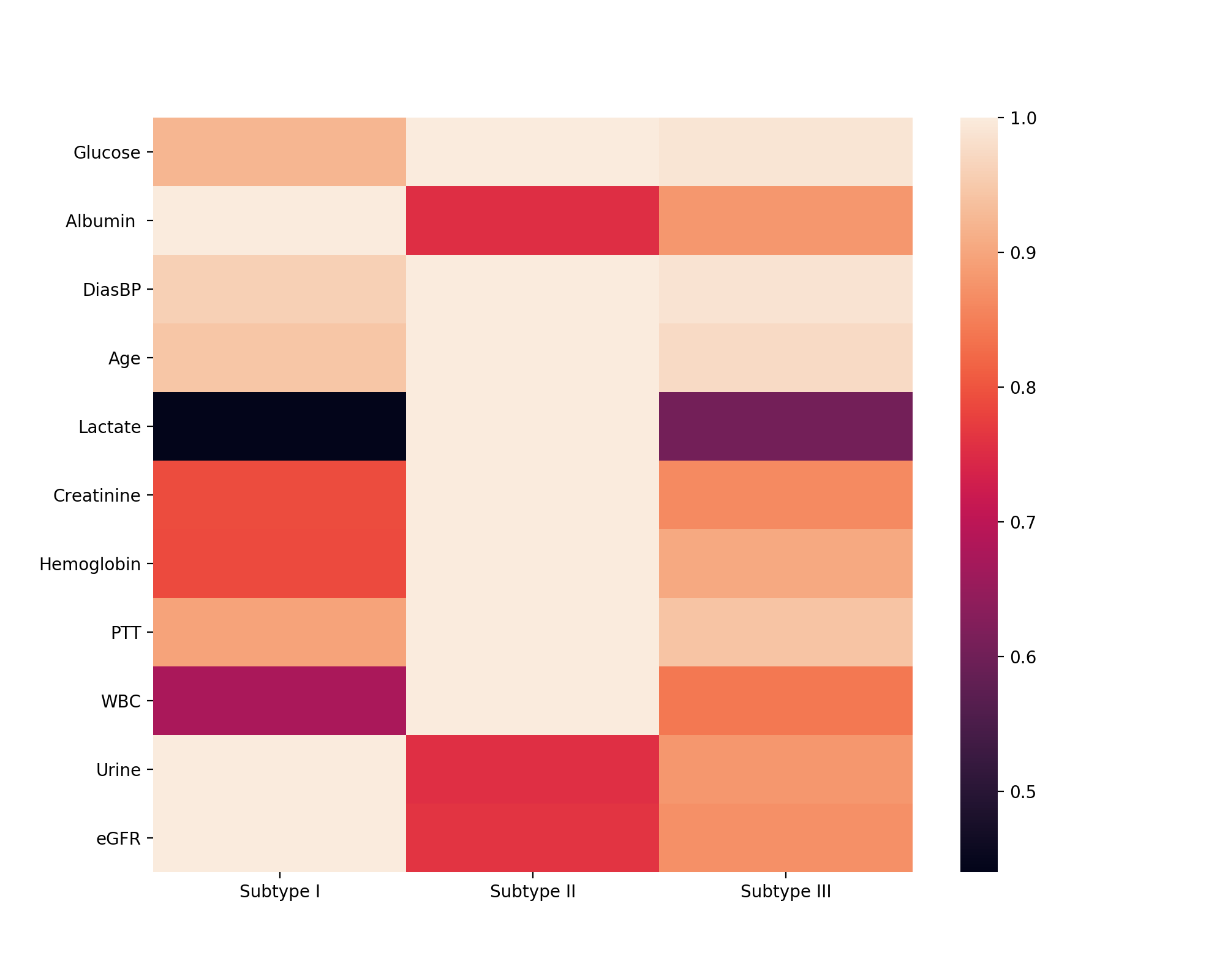}
	\caption{The illustration of heatmap in terms of significant continuous variables in each subtypes.}
	\label{FIG:heatmap}
\end{figure}

We further demonstrate the values of these features in different sub-phenotypes in Figure \ref{FIG:heatmap}. From the figure we can see that sub-phenotype I is with mild kidney dysfunction and sub-phenotype II is with severe kidney dysfunction, while sub-phenotype III is in between. 

\begin{table}
\centering
\caption{The results of statistical analyses on discrete variables (Number(Percentage), $^a$Chi-square test)} 
\label{tbl:table_Discrete}
{\small
\begin{tabular}{c|c|c|c|c|c }  
\hline\hline
      & Subtype I & Subtype II & Subtype III  & Unadjusted  & Adjusted\\
      & (N=4553)  & (N=672) & (N=2432) & P-value & ANCOVA$^*$\\
\hline\hline 
Male & 2928(64.31\%) & 310(46.13\%) & 1334(54.85\%) & 0.520$^a$ & 0.512\\
Female & 1625(35.69\%) & 362(53.87\%) & 1098(45.15\%) && \\
\hline 
Ethnicity\_White & 924(20.29\%)  & 94(13.99\%) &	603(24.79\%) && \\
Ethnicity\_Black	& 2514(55.22\%)  & 463(68.9\%)	&1315(54.07\%) & 0.567$^a$ & 0.602\\
Ethnicity\_Asian	&674(14.8\%)	&72(10.71\%)	&408(16.78\%) &&\\
Ethnicity\_others &441(9.69\%)	& 43(6.4\%)	& 106(4.39\%)&&\\
\hline
Diuretics	&601(13.2\%)	&171(25.44\%)	&406(16.69\%)	& 0.542$^a$ & 0.432\\
Nsaid	&555(12.19\%)	&167(24.85\%)	&399(16.41\%)	& 0.482$^a$  & 0.543\\
Angiotensin	&641(14.08\%)	&174(25.89\%)	&418(17.19\%)	& 0.540$^a$ & 0.603\\
CHF	&2739(60.16\%)	&441(65.63\%)	&1532(62.99\%)	& 0.534$^a$ & 0.624\\
PV:	&677(14.87\%)	&115(17.11\%)	&455(18.71\%)	&0.591$^a$ & 0.577\\
Hypertension	&2759(60.6\%)	&405(60.27\%)	&1433(58.92\%)	&0.589$^a$ & 0.612\\
Diabetes	&1409(30.95\%)	&298(44.35\%)	&1081(44.45\%)	& 0.457$^a$ & 0.521\\
LD	&602(13.22\%)	&103(15.33\%)	&341(14.02\%)	&0.591$^a$ & 0.614\\
Mi 	& 567(12.45\%)	&99(14.73\%)	&364(14.97\%)	& 0.063$^a$ & 0.145\\
Cad	& 1632(35.84\%)	&321(47.76\%)	&1210(49.75\%)	&0.465$^a$ & 0.513\\
Cirrhosis	&410(9.01\%)	&81(12.05\%)	&291(11.97\%)	& 0.563$^a$ & 0.633\\
Jaundice	&201(4.41\%)	&46(6.85\%)	&149(6.13\%)	&0.313$^a$ & 0.453\\
\hline\hline
\end{tabular}}

\vspace{0.5em}
$^*$ANCOVA was performed to adjust significant in terms of age variable. CHF: Congestive Heart Failure; PV: Peripheral Vascular; LD: Liver Disease; CAD: Coronary Artery Disease; MI: Myocardial Infarction.
\end{table}

\begin{sidewaystable}
\centering
\caption{The results of statistical analyses on continuous variables (Mean (Standard Deviation), $^b$One-way ANOVA test, $^c$Kruskal-Wallis H-test)} 
\label{tbl:table_Continuous}
{\small
\begin{tabular}{c|c|c|c|c|c}
\hline\hline
      & Subtype I & Subtype II & Subtype III & Unadjusted  & Adjusted  \\
      & (N=4553)  & (N=672) & (N=2432) & P-value & ANCOVA $^*$ P-value\\
\hline\hline
Glucose	&134.32(40.66)	&145.56(46.67)	&144.22(46.67)	&$<$0.001$^b$ (I vs II, III) & $<$0.001\\
Albumin	&3.99(0.52)	&3.01(0.64)	&3.51(0.51)	&$<$0.001$^c$ (I vs II, III, II vs III)  & $<$0.001 \\
AST 	&82.28(15.00)	&85.54(20.83)	&83.59(19.13)	&0.561$^b$ & 0.665\\
Bilirubin	&1.41(2.91)	&4.87(5.61)	&4.68(4.97)	&0.538$^c$ & 0.672\\
DiasBP 	&58.64(12.24)	&61.13(12.53)	&60.45(12.62)	&0.004$^b$ (I vs II, III) & 0.165\\
Age	&63.03(17.25)	&66.81(10.43)	&65.07(11.32)	&$<$0.001$^b$ (I vs II, III) &  - - - \\
Lactate	&2.16(1.05)	&4.91(1.58)	&2.97(1.38)	&0.012$^c$ (I vs II, III, II vs III) & 0.025\\
pH	&7.38(0.06)	&7.37(0.07)	&7.36(0.08)	&0.416$^c$ & 0.565\\ 
HeartRate	&87.22 (17.09)	&90.65(16.26)	&86.12(15.09)	&0.564$^b$ & 0.597\\
MeanBP 	&76.09(13.25)	&78.46(13.67)	&79.02(11.40)	&0.500$^b$ & 0.776\\
RespRate 	&18.08(4.44)	&20.26(4.75)	&19.19(4.01)	&0.403$^b$ & 0.465\\
SpO2	&96.37(1.97)	&96.27(2.16)	&97.23(2.13)	&0.481$^c$ & 0.374\\
SysBP	&115.67(15.94)	&120.22(18.11)	&120.43(17.63)	&0.432$^c$ & 0.254\\
Temp 	&36.85(0.62)	&36.82 (0.66)	&36.82(0.62)	&0.231$^b$ & 0.432\\
Bicarbonate	&23.87(4.16)	&24.70(4.83)	&24.51(4.60)	&0.542$^c$ & 0.654\\
BUN 	&28.66(22.74)	&28.65(24.77)	&27.66(21.34)	&0.501$^c$ & 0.776\\
Calcium	&8.36(0.73)	&8.40(0.74)	&8.78(0.70)	&0.378$^b$ & 0.443\\
Chloride	&105.19(5.45)	&102.22(5.90)	&103.38(5.62)	&0.3423$^b$ & 0.665\\
Creatinine	&1.55(0.34)	&1.96(0.49)	&1.69(0.32)	&$<$0.001$^c$ (I vs II, III, II vs III) & $<$0.001 \\
Hemoglobin	&13.55(1.76)	&17.18(1.55)	&15.53(1.91)	&$<$0.001$^b$(I vs II, III, II vs III)& 0.101\\
INR	&1.47(0.72)	&1.54(1.04)	&1.47(0.94)	&0.334$^c$ & 0.554\\
Platelet	&242.08(43.63)	&384.96(115.46)	&265.31(44.64)	&0.521$^b$ & 0.654\\
Potassium	&4.24(0.56)	&4.25(0.56)	&4.22(0.54)	&0.443$^c$ & 0.556\\
PT 	&15.45(5.76)	&17.30(7.45)	&15.55(6.38)	&0.346$^c$ &0.564\\ 
PTT 	&35.12(18.55)	&39.24(14.42)	&36.94(17.18)	&$<$0.001$^c$ (II vs III, I) &0.201\\
WBC 	&10.59(8.72)	&15.71(7.97)	&13.23(5.14)	&$<$0.001$^c$ (I vs II, III, II vs III) &$<$0.001\\
Urine	&1.35(0.24)	&1.02(0.25)	&1.19(0.25)	&$<$0.001$^c$ (I vs II, III, II vs III) &$<$0.001\\
eGFR	&107.65(54.98)	&82.19(55.92)	&93.97(56.53)	&$<$0.001$^c$ (I vs II, III, II vs III) &$<$0.001\\
\hline\hline
\end{tabular}}

\vspace{0.5em}
$^*$ANCOVA was performed to adjust significant in terms of age variable. AST: Aspartate Aminotransferase in blood; DiasBP: Diastolic Blood Pressure; MeanBP: Mean arterial Blood Pressure; RespRate: Respiration Rate; SysBP: Systolic Blood Pressure; Temp: Temperature; BUN: Blood Urea Nitrogen; INR: International Normalized Ratio; PT: Prothrombin Time; PTT: Partial Thromboplastin Time; WBC: White Blood Count; eGFR: estimated Glomerular Filtration Rate.
\end{sidewaystable}

In order to further validate our observations, we checked the AKI severity of those patients in different sub-phenotypes. We used the AKI staging criteria from KDIGO as we introduced in Section \ref{sec:dataset}. We found that sub-phenotype I is mainly with stage I, sub-phenotype II is mainly associated with stage III AKI, while sub-phenotype III is mainly associated with stage II AKI. The compositions of the three sub-phenotypes with respect to the AKI stages are summarized in Table~\ref{tbl:table_stage_subtype}.

\begin{table}
\centering
\caption{The statistical results for AKI stages and subtypes} 
\label{tbl:table_stage_subtype}
{\small
\begin{tabular}{c|c|c|c}
\hline\hline
      & Stage\_1 & Stage\_2 & Stage\_3\\
\hline
Subtype I	(N=4553)& 3236(71.07\%)	& 957(21.02\%)	&360(7.9\%) \\
\hline
Subtype II	(N=672)& 75(11.16\%)	& 171(25.45\%)	&426(63.39\%)\\
\hline
Subtype III	(N=2432)& 386(15.87\%)	& 1609(66.16\%)	&437(17.97\%)\\
\hline\hline
\end{tabular}}
\end{table}

\section{Conclusions and Future Directions}
In this paper we propose a data-driven approach for identification of AKI sub-phenotypes. Our approach is composed of three steps. In the first step we develop a memory network based architecture to predict the AKI risk by integrating both the structured and unstructured information in patient EHRs. In the second step we will perform clustering based on the patient representations derived from the first step. In the third step we identify important features that are significantly different across the different clusters and use them to interpret the clusters. On the MIMIC III data set, we identified three AKI sub-phenotypes, and they correlate well with the three stages of AKI very well.

There are two main points in terms of the clinical significance of this study. In particular, firstly, the experimental setting on the proposed model is based on prediction for a patient whether or not suffering from AKI in the next 7 days. The experimental results show that the proposed model has the potential to obtain a better representation. One of the clinical implications of this study is that,  for a new patient admitted in ICU, the proposed model can make a prediction with high accuracy for the patient whether or not suffering from AKI. Secondly, since the etiology of AKI is complex, it might be not enough to discover all the subtypes using known rules and some important factors. The proposed model used all raw data to obtain patient representations and discover subtypes, which has the potential to provide some important information assisting clinicians to understand the patient conditions (e.g., the severity of AKI in the next 7 days)  in time and take appropriate actions.

For the interpretation step, multiple-testing adjustment was applied to the p-value in our reported results. Note that, the clusters were trained using structured and unstructured data. The variables in reported statistical results are from structured data, which is a part of data used for training clusters. The p-values of these variables may be not very rigorous hypothesis testing.


In the future, our proposed approach can be improved from the following aspects
(1) The proposed method is completely data-driven. We can consider how to combine AKI domain knowledge in the model building process.
(2) We can add components such as attention mechanism to enhance the model interpretability.
(3) {We will consider more important variable categories (e.g., interventions, transfusions, diagnostics)\citep{koyner2018development,tomavsev2019clinically} and replicate the identified sub-phenotypes on more data sets.}

\section*{References}
\bibliographystyle{unsrtnat}
\bibliography{references}

\end{document}